\documentclass{scrartcl}
\usepackage[utf8]{inputenc}
\usepackage[T1]{fontenc}
\usepackage{graphicx}
\usepackage{grffile}
\usepackage{longtable}
\usepackage{wrapfig}
\usepackage{rotating}
\usepackage[normalem]{ulem}
\usepackage{amsmath}
\usepackage{textcomp}
\usepackage{amssymb}
\usepackage{capt-of}
\usepackage{hyperref}
\usepackage{subcaption}
\usepackage{footnote}
\usepackage{floatpag}
\usepackage{float}
\usepackage{setspace}
\usepackage{geometry}
\usepackage{tikz}
\usepackage{bm}
\usepackage{bbm}
\usepackage{wrapfig}
\usepackage{makeidx}
\usepackage{amsmath}
\usepackage{amsthm}
\usepackage{amsfonts}
\usepackage{amssymb}
\usepackage{cases}
\usepackage{pdfpages}
\usepackage{algorithm}
\usepackage{algpseudocode} 
\usepackage{todonotes}
\usepackage{natbib}
\makeindex
\usepackage{accents} 
\DeclareMathAlphabet{\mathpzc}{OT1}{pzc}{m}{it} 
\numberwithin{equation}{section}
\theoremstyle{definition}
\newtheorem{rem}{Remark}

\newcommand{\norm}[1]{\ensuremath{\left|\left|#1\right|\right|}}
\usepackage{mathtools}

\DeclarePairedDelimiterX{\inp}[2]{\langle}{\rangle}{#1, #2}

\DeclareMathOperator*{\argmin}{arg\,min}

\allowdisplaybreaks
\author{Renato Budinich\thanks{R. Budinich is with the Fraunhofer SCS,
Nordostpark 93, 90411 N\"urnberg, Germany, e-mail: renato.budinich@scs.fraunhofer.de}  \, \& Gerlind Plonka\thanks{G. Plonka is  with the Institute for Numerical and Applied Mathematics, University of G\"ottingen, Lotzestr. 16--18, 37083 G\"ottingen, Germany, email:plonka@math.uni-goettingen.de}}
\date{\today}
\title{A Tree-based Dictionary Learning Framework}
\hypersetup{
 pdfauthor={Renato Budinich \& Gerlind Plonka},
 pdftitle={A Tree-based Dictionary Learning Framework},
 pdflang={English}}
\begin{document}

\maketitle

\begin{abstract}
\label{sec:orgd8de846}
We propose a new outline for  adaptive dictionary learning methods for sparse encoding based on a hierarchical clustering of the training data.

Through recursive application of a clustering method the data is organized into a binary partition tree representing  a multiscale structure.
The dictionary atoms are defined adaptively based on the data clusters in the partition tree.
This approach can be interpreted as a generalization of a discrete Haar wavelet transform.
 Furthermore, any  prior knowledge on the wanted structure of the dictionary elements can be simply incorporated.
 The computational complexity of our proposed  algorithm depends on  the employed  clustering method  and on the chosen similarity measure between data points.
Thanks to the multiscale properties of the partition tree, our dictionary is structured: when using Orthogonal Matching Pursuit to reconstruct patches from a natural image, dictionary atoms corresponding to nodes being closer to the root node in the tree have a tendency to be used with greater coefficients.
\end{abstract}


\section{Introduction}
\label{sec:org27cace7}
In many applications one is interested in sparsely approximating a set of $N$ $n$-dimensional data points $Y_{j}$, columns 
of an $n \times N$ real matrix $\mathbf Y = (Y_{1}, \ldots, Y_{N})$. Assuming that the data can be efficiently represented in a transformed 
domain, given by applying a linear transform ${\mathbf D} \in {\mathbb R}^{n\times K}$, one is interested in  solving the sparse coding problem
\begin{align}
\label{eq:sparsecodingprob}
\min_{{\mathbf X} \in \mathbb{R}^{K \times N}} &\norm{{\mathbf Y} - {\mathbf D}{\mathbf X}} \, , \qquad \text{where} \quad 
\norm{X_{j}}_0 \leq S \quad \forall \, j=1,\ldots,N  \, ,
\end{align}
where $S \in \mathbb{R}$ is a parameter called \emph{sparsity}, ${X}_{j}$ is the $j$-th column of the encoding matrix ${\mathbf X}=(X_{1}, \ldots , X_{N}) \in {\mathbb R}^{K \times N}$ and $\norm{\cdot}_0$ is the so-called $0$-norm which is defined as the number of non-zero components of a vector (and is not really a norm). 
The $j$-th column of the \emph{encoding matrix} ${\mathbf X}$ gives the coefficients used in the linear combination of columns of ${\mathbf D}$ (which are termed \emph{atoms} of the \emph{dictionary}) to approximate the $j$-th column $Y_{j}$ of ${\mathbf Y}$. 
How well the data $Y_{j}$ can indeed  be approximated by ${\mathbf D} X_{j}$ with an $S$-sparse vector $X_{j}$ is of course dependent on ${\mathbf Y}$ and on the choice of ${\mathbf D}$.

The sparse coding problem in (\ref{eq:sparsecodingprob}) is NP-hard (see \cite{natarajan1995sparse}) and thus one can only hope to find an approximate minimizer ${\mathbf X}$. 
Within the last years a multitude of methods has been proposed to find approximated solutions to problem \eqref{eq:sparsecodingprob}. 
Most of these are greedy algorithms  that sequentially  select the $S$ dictionary atoms to approximate the columns $Y_{j}$ of ${\mathbf Y}$, as e.g.
Orthogonal Matching Pursuit (OMP)  or the Iterative Thresholding method by \cite{blumensath2008iterative}. Many approaches replace the $0$-norm in  (\ref{eq:sparsecodingprob}) by the $1$-norm to obtain a convex minimization problem that can in turn be solved efficiently, see e.g. \cite{beck09,chambolle11} and Basis Pursuit methods, see e.g.\ \cite{pati1993orthogonal,davies97, tropp04}.
For specific dictionary matrices exact solvers exists, see e.g.\ \cite{dragotti2014sparse} for ${\mathbf D} =[{\mathbf I}, {\mathbf F}]$ with 
${\mathbf I}$ the identity and ${\mathbf F}$ the Fourier matrix.

Finding a dictionary matrix ${\mathbf D}$ that admits the most efficient representation of the given data set ${\mathbf Y}$ is even more delicate.
The often considered synthesis \emph{dictionary learning} problem consists  in finding both the optimal transformation  ${\mathbf D}$ and the sparse coding matrix ${\mathbf X}$,
\begin{align}
\label{eq:dictlearnprob}
\min_{{\mathbf D} \in \mathbb{R}^{n \times K}, {\mathbf X} \in \mathbb{R}^{K \times N}} &\norm{{\mathbf Y} - {\mathbf D}{\mathbf X}} \qquad \text{where}  \quad \norm{X_{j}}_0 \leq S  \quad  \forall \, j=1,\ldots,N \,.
\end{align}
In this problem (which is also also NP-hard, see \cite{tillmann2015computational}) one is supposing that there exists 
an approximate factorization ${\mathbf D}{\mathbf X}$ of the matrix ${\mathbf Y}$ where ${\mathbf X}$ is (column-wise) sparse.
 A well-known method to tackle (\ref{eq:dictlearnprob}) is the K-SVD algorithm by \cite{aharon2006img}. 
 Many improvements and modifications of the K-SVD have been proposed within the last years, see e.g.\ \cite{ophir2011,nguyen2012,rubin2013,eksio2014}.
Another state-of the art approach for unsupervised dictionary learning is based on matrix factorization, see e.g. \cite{mairal2010}.

The models (\ref{eq:sparsecodingprob}) and (\ref{eq:dictlearnprob}) both implicitly assume that the given training data points $Y_{j}$ are vectors. 
However, in many applications the data already possesses a multidimensional spatial structure, which is not leveraged when the data points are vectorized into the columns of the matrix ${\mathbf Y}$. In the last years there have been attempts  to propose other dictionary learning methods, which on the one hand try to take the structure of the data into account and on the other hand  impose  further structure of the dictionary matrix in order to come up with more efficient dictionary learning algorithms for special applications, see e.g.\ \cite{yanke16,cai14,liu17,liu2018sparse}.

\medskip 
In this paper, we want to propose a general dictionary learning approach, which is based on organizing the training data into a binary tree corresponding to a hierarchical clustering, thereby providing a multiscale structure that we leverage to construct the dictionary atoms of ${\mathbf D}$. 
In particular, we completely separate the sparse coding problem (\ref{eq:sparsecodingprob}) and the problem of fixing the dictionary ${\mathbf D}$. 
 Our technique generalizes ideas in \cite{zeng2015dictionary} and \cite{liu2018sparse}, and particularly shows the connection to an adaptive multi-scale structure that can be interpreted as a data-driven generalized Haar wavelet transform.
\medskip

This idea significantly differs from earlier approaches using tree-structured dictionaries as e.g.\ in \cite{jenat2011,maza2013,shen2015} as well as from dictionary learning using wavelet frames \cite{ophir2011,sulam2016}.
\cite{jenat2011} aim at solving (\ref{eq:dictlearnprob}) with the assumption that the dictionary possesses a tree structure, where each dictionary element is identified with a tree knot. In \cite{maza2013}, a tree K-SVD is proposed, where several dictionaries are introduced, each dictionary is learned from a subset of residuals  of the previous level  using K-SVD. Similarly, \cite{shen2015} proposed a multi-level discriminative dictionary learning  method based on  several learned  hierarchical discriminative dictionaries.

The approaches in \cite{ophir2011,sulam2016} are based on  learning patch-based dictionaries (using K-SVD) in the analysis domain of the Wavelet transform, where \cite{sulam2016} particularly aims at dictionary learning in higher dimensions.

\medskip

\noindent 
Our dictionary learning process consists of two steps: the computation of a binary \emph{partition tree} which provides  a hierarchical adaptive clustering of the training data, and the determination of the dictionary elements from the partition tree. The partition tree is computed by means of recursive application of a two-way clustering method: depending on the type of the data, its structure and on computation speed requirements, one can choose a clustering method that is most appropriate. To this purpose it is possible to leverage certain pre-defined structure of the dictionary elements, as e.g.\  tensor product structure as in \cite{zeng2015dictionary} or rank conditions as proposed in \cite{liu2018sparse}.

In order to determine the dictionary elements from the partition tree we  propose  a procedure that can be interpreted as a generalization of the  Haar wavelet transform. 
To illustrate this analogy, we will show that the classical Haar wavelet transform can be transferred to a binary tree construction from bottom to top, the usual ``local to global'' approach. Due to its linearity and invertibility, this is however equivalent to a top to bottom construction, making it ``global to local''; this second approach is what we use in our method. This analogy allows us to see how our method constructs a multi-scale analysis of the data, much as the Haar wavelet transform does. The difference is that our method is adaptive, meaning that the tree is determined by the structure of the data, unlike the Haar wavelet tree which depends only on the number of data points. The multiscale property of the tree is reflected in our dictionary, with atoms higher up in the tree encoding differences at a lower resolution scale: we notice in fact that OMP has a tendency to use these atoms with greater coefficients.

\noindent Having found the dictionary matrix ${\mathbf D}$ from the clusters in the binary tree, we still need to solve the sparse coding problem (\ref{eq:sparsecodingprob}).
For our application we will use OMP to sparsely code the data. We compare our method with K-SVD in various natural image reconstruction tasks: it usually performs slightly worse in terms of quality of the reconstruction but is faster especially for growing number of data points. This is to be expected since, when using Lloyd's algorithm for K-means, our algorithm has linear complexity.

\medskip
The structure of this paper is as follows. In Section \ref{sec2} we extensively describe the proposed procedure for dictionary learning.
We start with the construction of the binary partition tree  in Section \ref{sec:part} and show in Section \ref{sect:dict} how to extract the dictionary atoms from the partition tree. Section \ref{sec:toy} illustrates the two steps for dictionary learning in a toy example. In Section \ref{sec:Haar}, we present the connection of our dictionary construction with an adaptive Haar wavelet transform. Section \ref{sec:alg} is concerned with some algorithmic aspects of the dictionary learning procedure. In Section \ref{sec:num} we present some application results for various reconstruction tasks comparing our method to K-SVD.

\section{Tree-based dictionary learning framework}
\label{sec2}

Differently from other dictionary learning methods, where the dictionary matrix ${\mathbf D}$ and the sparse coding matrix ${\mathbf X}$ are optimized simultaneously, our proposed method concerns itself only with learning  ${\mathbf D}$; a sparse coding method such as OMP must be employed in a second, separate step. 

Assume that  we are given set of data ${\mathbf Y} = \{ Y_{1}, \ldots , Y_{N} \}$, where  all $Y_{j}$ have the same known data structure.
The $Y_{j}$ can for example be vectors $Y_{j} \in {\mathbb R}^{n}$,  image patches $Y_{j} \in {\mathbb R}^{m_{1} \times m_{2}}$, tensors, or more generally have a  finite graph structure.
For our dictionary learning method, the $Y_{j}$ can be any type of data, as long as:
\begin{itemize}
\item we have a meaningful clustering method for it, which should ideally separate the data according to salient features;
\item we can take linear combinations of the samples.
\end{itemize}

If the $Y_{j}$ have a graph structure, we thus will need to define  addition and multiplication with scalars on graphs suitably.
For simplicity, in this paper we will thus ask that the samples live in a normed vector space $V$. 

 The dictionary  learning process itself consists of two parts:
\begin{enumerate}
\item  computation of a binary partition tree which gives a hierarchical clustering of the training data,
\item  determination of dictionary elements from the partition tree.
\end{enumerate}
Within the next two subsections we will introduce notations and describe these two steps in detail.

\subsection{Construction of  the partition tree}
\label{sec:part}
We assume that each data sample $Y_{j}$ can be uniquely identified by its index $j \in \{1, \ldots , N\}$. We want to construct a binary partition tree $T$ whose nodes are associated to subsets of the index set $\{ 1, \ldots , N\}$, i.e., each node must correspond to a unique subset of the training data. We will interchangeably identify the nodes with the subset of indexes or of data points - this should be clear from the context and won't be  source of ambiguity.
Let the root node be 
$${\mathcal N}_{0,0} := \{1, \ldots , N \}. $$ 
In general,  $\mathcal{N}_{\ell,k }$ is the node at level $\ell$ that has $\mathcal{N}_{\ell+1, 2k}$ and $\mathcal{N}_{\ell +1, 2k +1}$ as children nodes. For a  binary tree with a \emph{complete} level $\ell$  we have $2^{\ell}$ nodes in this level, i.e., there will be nodes $\mathcal{N}_{\ell,k}$ for all $k \in \{ 0, \ldots , 2^{\ell}-1 \}$. If the level is not complete, there will be nodes $\mathcal{N}_{\ell,k}$ only for certain values of $k$.
We call this tree the \emph{partition} tree because for each (non-leaf) node ${\mathcal N}_{\ell,k}$ of the tree,  the two children nodes satisfy the properties
\begin{eqnarray}
\nonumber
  {\mathcal N}_{\ell+1,2k} \cup {\mathcal N}_{\ell+1,2k+1} &=& {\mathcal N}_{\ell,k} \,\,,\\
  \label{set}
  {\mathcal N}_{\ell+1,2k}\,,\,{\mathcal N}_{\ell+1,2k+1} \,\, &\neq& \emptyset \,\, \mbox{and} \\
  \nonumber
  {\mathcal N}_{\ell+1,2k} \cap {\mathcal N}_{\ell+1,2k+1} &=& \emptyset. 
\end{eqnarray}
The tree $T$ is generated  by recursive application  of a clustering  method  to partition a given subset of the data into two subsets. The tree $T$ obtained in this way need not to be complete (i.e., not all leaf nodes will in general be at the same level), and the number of elements in the subsets ${\mathcal N}_{\ell+1,2k}$ and ${\mathcal N}_{\ell+1,2k+1}$  need not to be the same.  Thus, we will need some rule  to decide whether  a node set ${\mathcal N}_{\ell,k}$  will be partitioned into  two further subsets or not; we will discuss this also in Section \ref{sec:alg}.
\medskip

In order to  obtain a meaningful partition tree  that leads to a good dictionary, we need  to choose an appropriate  clustering procedure. If the data  (and thus also the dictionary elements that we want to construct) should have a certain special structure, as e.g., block circulant or block Toeplitz matrices, then this could be leveraged; a dimension reduction step like PCA or 2DPCA (see \cite{kong2005generalized}) could also be used to accelerate computation.

In our numerical experiments in Section \ref{sec:num} we will compare K-means, K-maxoids (see \cite{bauckhage2015k}), both with $K=2$, and Spectral Clustering. 2-maxoids and especially 2-means are faster due to the lower computational complexity of the algorithms. From a practical point of view, the main difference between the two is that while 2-means offers as representatives of the clusters the sample average of data points therein contained, 2-maxoids gives as representative a particular data point. Spectral Clustering has the theoretical advantage that it can be applied on a data-graph built in any way from the data: one isn't restricted to the Euclidean distance but can cluster the data based on any type of similarity measure between the data points.

In the remainder of the section, 
we present some possible strategies for the construction of a binary tree in order to illustrate that there is a large variety to connect clustering methods with a dimensionality reduction as a preprocessing step.

\begin{enumerate}
\item \textbf{2-means clustering and FIFO procedure without preprocessing of training data.}
Assume that we have a set of $(n \times m)$ image patches as training data $\{Y_{1}, \ldots , Y_{N}\}$. We fix the first root node of indices ${\mathcal N}_{0,0} := \{1,2, \ldots , N\}$.
We  create the tree using 2-means clustering and the \texttt{FIFO} (first in first out) queue  with parameters \texttt{mincard} and $\epsilon$; see Section \ref{sec:alg}. This means, further branching of a subset in the tree will be performed if the cardinality of a node is above a predetermined constant \texttt{mincard} $\in {\mathbb N}$ and the clustering minimization function is above the predetermined threshold $\epsilon >0$. \\
Initially 2-means is applied to the full set of training patches corresponding to the node $\mathcal{N}_{0,0}$, separating 
a subset of $\{1, \ldots , N\}$ from the rest. Observe that the two obtained subsets usually do not have the same size.
We repeat the procedure for the subsets to construct the partition tree. In this process, the number of nodes of the tree is not fixed in advance.

\item \textbf{2-means clustering and FIFO queue applied to one-dimensional features of training data.}
Assume again that we have a set of $(n \times m)$ image patches as training data $\{Y_{1}, \ldots , Y_{N}\}$. This time, we 
first use a dimensionality reduction procedure before employing the $2$-means algorithm.
For the first partition, we compute the centroid ${A}_{0,0}:= \frac{1}{N} \sum_{j=1}^{N} Y_{j}$ and evaluate the spectral norms of the difference matrices 
$s_{j} := \| {A}_{0,0} - Y_{j} \|_{2}$ for $j=1, \ldots, N$, thereby reducing the $Y_{j}$ to a one-dimensional feature to measure the deviation of $Y_{j}$ from $A_{0,0}$. Let 
$$ s_{r_{1}} \le s_{r_{2}} \le \ldots \le s_{r_{N}} $$
be the obtained ordered feature numbers of training data. In this case, the 2-means algorithm reduces to the minimization problem 
\begin{equation}\label{2m} \hat \mu := \argmin_{1 \le \mu \le N-1}  \left[
\sum_{n=1}^{\mu} \left( s_{r_{n}}- \frac{1}{\mu} \sum_{\nu=1}^{\mu} s_{r_{\nu}}\right)^{2} + \sum_{n=\mu+1}^{N}
\left( s_{r_{n}} - \frac{1}{N-\mu
} \sum_{\nu= \mu+1}^{N} s_{r_{\nu}}\right)^{2} \right] \,\,,
\end{equation}
see \cite{budinich18}, Theorem 1.7.2. In other words, the 2-means algorithm provides a uniquely defined optimal solution,
and we obtain the partition into the two index sets $\{r_1, \ldots, r_{\hat{\mu}}\}$ and $\{r_{\hat{\mu}+1}, \ldots, r_N\}$.
We repeat the procedure for the subsets to construct the partition tree using predetermined constants \texttt{mincard} and $\epsilon$.

\item \textbf{Spectral clustering and PRIORITY queue applied to low-rank approximations of training data.}
Consider the given training set $\{Y_{1}, \ldots , Y_{N}\}$ of $(n \times m)$ image patches and compute in a preprocessing step rank-$r$ approximations
of all elements $Y_{1}, \ldots , Y_{N}$  using partial SVD. A similar approach has been applied in \cite{liu2018sparse} taking rank-$1$ approximations of 
noisy image patches $Y_{j}$.
Instead of using low-rank approximations of the training data, we can also apply a Fourier, DCT or wavelet transform to all $Y_{j}$ in a first step to obtain $M$-term approximation of $Y_{j}$, where only the $M$ largest coefficients in the Fourier/DCT/wavelet expansion of $Y_{j}$ are kept.
Then for the clustering step, instead of using 2-means, spectral clustering is used; this method leverages the spectral properties of the graph Laplacian matrix of the data similarity graph, see \cite{shi2000normalized,yan2009fast}.\\
Further, instead  of the FIFO queue procedure with parameters \texttt{mincard} and $\epsilon$,  we can apply the Priority queue procedure to construct the binary tree. Here, we fix an upper bound for the number of branchings $K$ in the binary tree in advance. Then, at each level of the tree, we first inspect all data subsets corresponding to the nodes in this level and start with branching the subset with highest variance first. 
The branching stops if either the cardinality of all subsets is not above \texttt{mincard} or the total number of branchings has reached $K$.
\end{enumerate}

\begin{rem}\label{rem:mary}
  Instead of a binary partition tree one may also think about constructing an  $m$-ary tree with $m>2$, by setting the clustering method to partition the data sets in $m$ subsets. In this case though the way the dictionary is built from the hierarchical clustering representation of the data set should change, see Remark \ref{rem:mtap}
\end{rem}

\subsection{Dictionary construction from the partition tree}
\label{sect:dict}

In this subsection we will describe how to extract the dictionary elements from the partition tree.
We will apply a multiscale procedure which is borrowed from the discrete  Haar wavelet transform but is here applied to our adaptive setting.
To extract the dictionary from the partition tree we have to make a choice for the \emph{representative} of each node: we will in general use the sample averages, and call these the representatives of the subset. The optimal way  to choose the representatives depends on the application, the clustering method and the  structure of the data; see below for some examples.

\noindent Once we have fixed the representatives for each node, we take as dictionary atoms
\begin{enumerate}
\item a first ``low-pass'' element, which is given by the representative of the root node;
\item for each node ${\mathcal N}_{\ell,k}$ in the partition tree which possesses two children nodes ${\mathcal N}_{\ell+1,2k}$ and ${\mathcal N}_{\ell+1,2k+1}$, the difference between the representatives of the children nodes.
\end{enumerate}

Let $A_{\ell,k}$ denote the representatives of the tree nodes, then we can define the normalized dictionary atoms as 
\begin{align}\label{D}
  \begin{split}
  \tilde{A}_{0,0} &= \frac{1}{N} \sum_{j=1}^{N} Y_{j}, \qquad A_{0,0} = \frac{\tilde{A}_{0,0}}{\|\tilde{A}_{0,0}\|}, \\
  \tilde{D}_{\ell,k} 
  &= A_{\ell+1,2k} - A_{\ell+1,2k+1}, \qquad     D_{\ell,k} = \frac{\tilde{A}_{\ell,k}}{\|\tilde{A}_{\ell,k}\|} \, ,
\end{split}
 \end{align}
 where we use the norm of the underlying vector space for normalization of the dictionary elements.
In case of image patches we will always use the Frobenius norm. We call the obtained dictionary the \textbf{Haar dictionary} and denote it with ${\mathbf D}^{H}$, i.e.
 \begin{align}
   \label{def:DH}
   \mathbf{D}^H := \left\{ A_{0,0}, D_{0,0}, D_{1,0}, D_{1,1,}, \ldots \right\}\,\,.
 \end{align}
 The choice of this name will become clear in Section \ref{sec:Haar}, where we show the connection to the discrete Haar wavelet transform.
Indeed, the obtained dictionary can be understood as an adaptive Haar wavelet frame.

As the representative of a node set of the tree we can simply take its centroid, which is  defined for the node $(\ell,k)$ as
\begin{align}\label{def:centroid}
A_{\ell,k} := \frac{1}{|\mathcal{N}_{\ell,k}|}\sum_{j \in \mathcal{N}_{\ell,k}} Y_j.
\end{align} 

However, we are not limited to choosing the centroids as \emph{representatives}  of a subset of training data to construct the Haar dictionary: in this regard there is a large variety of  possibilities, where in particular special dictionary structure can be incorporated. 
The choice of representative can be taken in accordance with the used procedure for tree construction in the last subsection, as e.g.\ the data  preprocessing method and the chosen clustering procedure.
We list some examples, how these representatives can be taken.

\begin{enumerate}
\item If  2-maxoids have been used as clustering method for the binary tree construction, then we obtain as outputs not only the partition but also two maxoids, which are particular data points belonging to each of these two subsets respectively. Thus, when using 2-maxoids, we can define $A_{\ell,k}$ as the maxoid of node $\mathcal{N}_{\ell,k}$. 
This procedure has been also used in our numerical experiments with $2$-maxoids in Section \ref{sec:num}.

\item We can take rank-$r$ approximations of the centroids in (\ref{def:centroid}). In this case the dictionary elements $D_{\ell,k}$ in (\ref{D}) have at most rank $2r$.

\item We can use $M$-term Fourier or wavelet expansions of the centroids as representatives of the subsets. (The wavelet transform can be taken according to the data structure and is independent from the adaptive Haar wavelet transform in (\ref{D}).)
The atom $D_{\ell,k}$ of $\mathbf{D}^H$ would then be the normalized difference of the two obtained $M$-term approximations $\hat{A}_{\ell+1,2k}$ and $\hat{A}_{\ell+1,2k+1}$. The obtained dictionary atoms would thus possess at most  $2M$ terms in the used Fourier/wavelet expansion.

\end{enumerate}

\begin{rem} 1. In \cite{liu2018sparse}, rank-1 approximations of the centroids have been used for constructing a dictionary $D$.
This construction provides dictionary elements of rank at most $2$ and is particularly suitable for noisy training patches.

2. By contrast, one can also take only the normalized representatives  corresponding to the leaves of the partition tree as dictionary atoms. 
We will call this the  \textbf{leaves dictionary} and denote it with $\mathbf{D}^L$, i.e.
 \begin{align}
   \label{def:DC}
   \mathbf{D}^L := \left\{ A_{\lambda_1}, A_{\lambda_1}, \ldots \right\}\,\,,
 \end{align}
 where $\lambda_1,\lambda_2,\ldots$ are the leaves of the partition tree.
An approach similar  to the leaves  dictionary construction has been also taken in \cite{zeng2015dictionary}, where the   representatives  corresponding to the leaves of the partition tree have been  rank-$d$ approximations of the centroids.
The atoms in the leaves dictionary  ${\mathbf D}^{L}$ corresponding to the lower nodes in the tree may potentially suffer from excessively high correlation, given that they represent clusters in close proximity of one another. It is known that high correlation between dictionary atoms is not ideal for sparse representation (see for example \cite{elad2010sparse}). Therefore we would advise to use the Haar dictionary $\mathbf{D}^H$ instead, especially for very deep trees.
\end{rem}

\subsection{A toy example for constructing a tree-based dictionary}
\label{sec:toy} 

To illustrate the construction of the partition tree and the determination of the dictionary we consider a toy example with $3 \times 3$ image patches.
Assume that we are given the  set of  training patches
\small
\begin{eqnarray*}
 \textstyle
Y_{1} \!\!\!\!&=&\!\!\!\!\left(\! \begin{array}{ccc} 1 & 0 & 0 \\ 1 & 2 & 0 \\ 0 & 1 & 3\end{array} \!\right)\!, \; 
Y_{2}=\left( \!\begin{array}{ccc} 1 & 0 & 0 \\ 1 & 2 & 0 \\ 0 & 1 & 5\end{array} \!\right)\!, \; 
Y_{3}=\left( \!\begin{array}{ccc} 1 & 0 & 0 \\ 1 & 1 & 0 \\ 1 & 0 & 0\end{array} \!\right)\!, \; 
Y_{4}=\left( \! \begin{array}{ccc} 2 & 0 & 0 \\ 5 & 5 & 0 \\ 2 & 7 & 5\end{array} \! \right)\!, \\
Y_{5}\!\!\!\! &=& \!\!\!\left(\! \begin{array}{ccc} 1 & 0 & 0 \\ 0 & 2 & 0 \\ 0 & 0 & 5\end{array} \!\right)\!, \, 
Y_{6}=\!\left( \!\begin{array}{ccc} 2 & 2 & 0 \\ 3 & 5 & 1 \\ 2 & 5 & 7\end{array} \!\right)\!, \, 
Y_{7}= \!\left( \!\begin{array}{ccc} 0 & 0 & 0 \\ 0 & 0 & 0 \\ 0 & 1 & 2\end{array} \!\right)\!, \, 
Y_{8}= \! \left(\! \begin{array}{ccc} 1 & 0 & 0 \\ 1 & 2 & 0 \\ 0 & 0 & 0\end{array} \!\right)\!.
\end{eqnarray*}
\normalsize

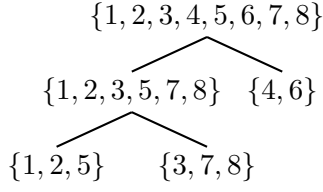
\begin{figure}[!h]
\begin{center}
\begin{tikzpicture}
\node[inner sep=2pt] (root) at (-4,0)
    {$\{1,2,3,4,5,6,7,8\}$};
\node[inner sep=2pt] (r0) at (-5,-1)
    {$\{1,2,3,5,7,8\}$};
\node[inner sep=2pt] (r1) at (-3,-1)
    {$\{4,6\}$};
    \draw[-,thick] (root.south) -- (r0.north);
    \draw[-,thick] (root.south) -- (r1.north);
    
\node[inner sep=2pt] (r00) at (-6,-2)
    {$\{1,2,5\}$};
\node[inner sep=2pt] (r01) at (-4,-2)
    {$\{3,7,8\}$};
    \draw[-,thick] (r0.south) -- (r00.north);
    \draw[-,thick] (r0.south) -- (r01.north);

\end{tikzpicture}
\end{center}
\caption{\small Partition tree obtained by applying the FIFO queue procedure with \texttt{mincard}$=3$, $\epsilon = 1$ and 2-means-clustering to the data.}
\label{fig-ex1}
\end{figure}

To construct the partition tree, we fix first the root node ${\mathcal N}_{0,0} := \{1,2,3,4,5,6,7,8\}$. We  create the tree using 2-means clustering and the \texttt{FIFO} queue procedure with parameters \texttt{mincard}$=3$ and $\epsilon = 1$; see Section \ref{sec:alg}. This means, further branching will only be performed if the cardinality of a node is above \texttt{mincard} and the clustering minimization function is above the threshold $\epsilon$. 
 Initially 2-means is applied to the full set of training patches corresponding to the node $\mathcal{N}_{0,0}$, separating patches $Y_4$ and $Y_6$ from the rest; then it is run on node  $\mathcal{N}_{1,0} = \{1,2,3,5,7,8\}$, splitting it into $\mathcal{N}_{2,0} = \{1,2,5\}$ and $\mathcal{N}_{2,1} = \{3,7,8\}$. The tree obtained is displayed in Figure \ref{fig-ex1}.
\medskip

\normalsize

For comparison, we also employ another approach to determine the partition tree to the same training data set, where we first use a dimensionality reduction procedure before employing the $2$-means algorithm.
For the first partition, we compute the  centroid ${A}_{0,0}:= \frac{1}{8} \sum_{j=1}^{8} Y_{j}$ and evaluate the spectral norms of the difference matrices 
$s_{j} := \| {A}_{0,0} - Y_{j} \|_{2}$ for $j=1, \ldots, 8$, thereby reducing the $Y_{j}$ to a one-dimensional feature. 

\begin{figure}[!h]
\begin{center}
\begin{tikzpicture}
\node[inner sep=2pt] (root) at (-4,0)
    {$\{1,2,3,4,5,6,7,8\}$};
\node[inner sep=2pt] (r0) at (-5,-1)
    {$\{1,2,3, 5,7,8\}$};
\node[inner sep=2pt] (r1) at (-3,-1)
    {$\{4,6\}$};
    \draw[-,thick] (root.south) -- (r0.north);
    \draw[-,thick] (root.south) -- (r1.north);
    
\node[inner sep=2pt] (r00) at (-6,-2)
    {$\{1,7\}$};
\node[inner sep=2pt] (r01) at (-4,-2)
    {$\{2, 3, 5, 8\}$};
    \draw[-,thick] (r0.south) -- (r00.north);
    \draw[-,thick] (r0.south) -- (r01.north);
  \node[inner sep=2pt] (r10) at (-5,-3)
    {$\{5,8\}$};
\node[inner sep=2pt] (r11) at (-3,-3)
    {$\{2, 3\}$};
    \draw[-,thick] (r01.south) -- (r10.north);
    \draw[-,thick] (r01.south) -- (r11.north);  

\end{tikzpicture}%
\end{center}
\caption{\small Partition tree obtained by applying by applying the FIFO queue procedure with \texttt{mincard}$=3$ and 2-means-clustering to the reduced data.}
\label{fig-ex2}
\end{figure}
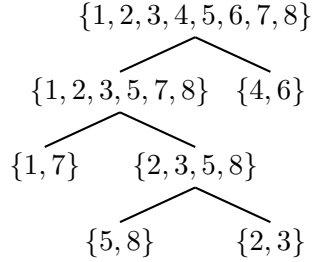

 Ordering the feature numbers of the training data we find here
$$s_{1} < s_{2} < s_{5}< s_{7}< s_{8}<s_{3}< s_{6}<s_{4}. $$
In this special case, the 2-means algorithm reduces to  the minimization problem (\ref{2m}) with $N=8$.
We obtain the partition into
${\mathcal N}_{1,0} = \{1,2,5,7,8,3\}$, ${\mathcal N}_{1,0} =\{6,4\}$. We proceed further in the same way for partitioning these subsets and obtain the tree in Figure \ref{fig-ex2}. Here we have applied a partition of a subsets as long as we have more than two entries in this set.

Thus, we indeed obtain a different partition tree, depending on different features used for clustering.

\noindent
Next,  we want to construct the dictionary from the partition tree.  We consider only the first partition tree in Figure \ref{fig-ex1}.
 As representatives of the $5$ nodes, we employ the
centroids (rounded to two digits):
{\small
\begin{eqnarray*}
{A}_{0,0} \!\!\!\!&=&\!\!\!\! \left(\! \begin{array}{ccc} 1.13 & 0.25 & 0 \\ 1.5 & 2.38 & 0.13 \\ 0.63 & 1.88 & 3.38 \end{array} \!\right), \; 
{A}_{1,0} \!=\!  \left( \!\begin{array}{ccc} 0.83 & 0 & 0 \\ 0.67 & 1.5 & 0 \\ 0.17 & 0.5 & 2.5 \end{array} \!\right), \;
{A}_{1,1} \!= \! \left(\! \begin{array}{ccc} 2 & 1 & 0 \\ 4 & 5 & 0.5 \\ 2 & 6 & 6 \end{array} \! \right), \\
  {A}_{2,0}\!\!\!\!&=&\!\!\!\! \left( \!\begin{array}{ccc} 1 & 0 & 0 \\ 0.67 & 2 & 0 \\ 0 & 0.67 & 4.3 \end{array}\! \right),  \; 
{A}_{2,1} = \left( \begin{array}{ccc} 0.67 & 0 & 0 \\ 0.67 & 1 & 0 \\ 0.33 & 0.33 & 0.67 \end{array} \right). 
\end{eqnarray*}
}
\normalsize
We thus obtain the Haar dictionary  ${\mathbf D}^{H}$ with $3$ elements 
$$
    \mathbf{D}^H := \left\{\frac{A_{0,0}}{\norm{A_{0,0}}_{F}}, \frac{A_{1,0} - A_{1,1}}{\norm{A_{1,0} - A_{1,1}}_{F}}, \frac{A_{2,0} - A_{2,1}}{\norm{A_{2,0} - A_{2,1}}_{F}} \right\} \, .
$$
By contrast, the leaves dictionary that uses beside the low-pass element $\frac{A_{0,0}}{\norm{A_{0,0}}_{F}}$ all normalized centroids of the leaves of the tree, is of the form 
$$
\mathbf{D}^L := \left\{\frac{A_{0,0}}{\norm{A_{0,0}}_{F}},\frac{A_{2,0}}{\norm{A_{2,0}}_{F}},\frac{A_{2,1}}{\norm{A_{2,1}}_{F}}, \frac{A_{1,1}}{\norm{A_{1,1}}_{F}}\right\} \, .
$$
Note that the leaves dictionary $\mathbf{D}^L$ has one element more than the Haar dictionary $\mathbf{D}^H$ - this is always the case.

\subsection{The Haar-dependency Tree}
\label{sec:Haar}

In this subsection we want to show the connection between the Haar wavelet dictionary and our tree-based dictionary construction. 
We start by recalling some basic facts about the  
discrete wavelet transform with  Haar low-pass filters $\{ \frac{1}{\sqrt{2}}, \, \frac{1}{\sqrt{2}} \}$  and associated high-pass filter
$\{ \frac{1}{\sqrt{2}}, \, -\frac{1}{\sqrt{2}} \}$,  
see for example \cite{damelin2012mathematics} or \cite{mallat2008wavelet}. 
Suppose we are given a digital signal ${\mathbf a}  \in \mathbb{R}^N$, for simplicity let $N = 2^L$ for some $L \in {\mathbb N}$ and denote the $N$ components of ${\mathbf a}_{L} := {\mathbf a}$ by $a_{L,0}, \ldots, a_{L,N-1}$. 
For $j= L-1,\ldots, 0$, we  define the recursive formulas for the so-called \emph{approximation} and \emph{detail} coefficients of the transform as
\begin{align}
\label{eq:haarsynthesis}
\begin{split}
a_{j,k} &= \frac{1}{\sqrt{2}}\big( a_{j+1,2k} + a_{j+1,2k+1}\big),  \\
d_{j,k} &= \frac{1}{\sqrt{2}}\big( a_{j+1,2k} - a_{j+1,2k+1}\big), 
\end{split} \qquad  k = 0,1, \ldots, 2^j - 1\,.
\end{align}
These formulas are known as  \emph{analysis formulas} and $j$ is known as the \emph{level} of the transform: the lower the level the coarser approximation the coefficients $a_{j,k}$ provide since they are an average of more samples. 
By applying the analysis formulas recursively $L$ times, the vector ${\mathbf a}_{L}  = (a_{L,0}, \ldots , a_{L,N-1})^{T}$  is linearly transformed into the vector
$(a_{0,0}, d_{0,0}, {\mathbf d}_{1}^{T}, {\mathbf d}_{2}^{T}, \ldots , {\mathbf d}_{L-1}^{T})^{T}$, where ${\mathbf d}_{j} := (d_{j,0}, \ldots ,$ $ d_{j,2^{j}-1})^{T}$ contains  the detail or wavelet coefficients of level $j$.

The  analysis formulas are easily inverted to obtain the \emph{reconstruction formulas}
\begin{align}
\label{eq:haarreconstruction}
  \begin{split}
  a_{j+1,2k} &= \frac{1}{\sqrt{2}}(a_{j,k} + d_{j,k}) ,\\
  a_{j+1,2k+1} &= \frac{1}{\sqrt{2}}(a_{j,k} - d_{j,k}).
\end{split}
\end{align}
The linear transform is hence invertible, and in fact even orthogonal.

\begin{figure}
\begin{center}
\begin{tikzpicture}
\node[inner sep=2pt] (a21) at (-4,0)
    {$a_{2,0}, d_{2,0}$};
\node[inner sep=2pt] (a31) at (-5,-1)
    {$a_{3,0}$};
\node[inner sep=2pt] (a32) at (-3,-1)
    {$a_{3,1}$};
\draw[-,thick] (a21.south) -- (a31.north);
\draw[-,thick] (a21.south) -- (a32.north);
\node[inner sep=2pt] (a22) at (-1,0)
    {$a_{2,1}, d_{2,1}$};
\node[inner sep=2pt] (a33) at (-2,-1)
    {$a_{3,2}$};
\node[inner sep=2pt] (a34) at (0,-1)
    {$a_{3,3}$};
\draw[-,thick] (a22.south) -- (a33.north);
\draw[-,thick] (a22.south) -- (a34.north);
\node[inner sep=2pt] (a23) at (2,0)
    {$a_{2,2}, d_{2,2}$};
\node[inner sep=2pt] (a35) at (1,-1)
    {$a_{3,4}$};
\node[inner sep=2pt] (a36) at (3,-1)
    {$a_{3,5}$};
\draw[-,thick] (a23.south) -- (a35.north);
\draw[-,thick] (a23.south) -- (a36.north);
\node[inner sep=2pt] (a24) at (5,0)
    {$a_{2,3}, d_{2,3}$};
\node[inner sep=2pt] (a37) at (4,-1)
    {$a_{3,6}$};
\node[inner sep=2pt] (a38) at (6,-1)
    {$a_{3,7}$};
\draw[-,thick] (a24.south) -- (a37.north);
\draw[-,thick] (a24.south) -- (a38.north);
\node[inner sep=2pt] (a11) at (-2.5,1.25)
    {$a_{1,0}, d_{1,0}$};
\draw[-,thick] (a11.south) -- (a21.north);
\draw[-,thick] (a11.south) -- (a22.north);
\node[inner sep=2pt] (a12) at (3.5,1.25)
    {$a_{1,1}, d_{1,1}$};
\draw[-,thick] (a12.south) -- (a23.north);
\draw[-,thick] (a12.south) -- (a24.north);
\node[inner sep=2pt] (a0) at (0.75,2.5)
    {$a_{0,0},d_{0,0}$};
\draw[-,thick] (a0.south) -- (a11.north);
\draw[-,thick] (a0.south) -- (a12.north);
\end{tikzpicture} \caption{The Haar dependency tree for \( N =8 \).\label{fig:haardeptree}}
\end{center}
\end{figure}
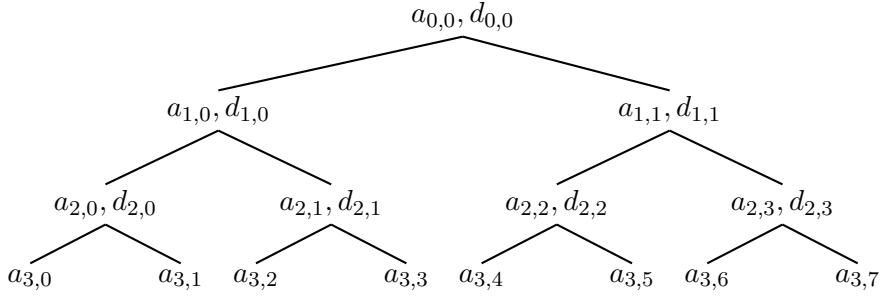

It is possible to represent the dependency between approximation coefficients in the analysis formulas by means of a binary tree (similarly to what is done in \cite{murtagh2007haar}), by  associating a node to each \( a_{j,k} \) which has two sons, \( a_{j+1,2k} \) and \( a_{j+1,2k+1} \). We start by identifying each of the original samples \( a_{L,k} \) with a leaf node, and subsequently for each level \( j=L-1,\ldots,0 \), and for each \( k=0,\ldots,2^j - 1 \), we add a node (corresponding to \( a_{j,k} \)) which has as sons the two coefficients at the previous levels from which it is computed, i.e. \( a_{j+1,2k} \) and \( a_{j+1,2k+1} \). If we apply the full \( L \) levels of the Haar wavelet transform we obtain a binary tree with root node \( a_{0,0} \); note that the concept of level of the tree and of the Haar wavelet transform here coincide, with the root node being at level \( 0 \) and the original samples at level \( L \).
In Figure \ref{fig:haardeptree} we show this tree for \( N=8 \): here we're labeling each node with the respective approximation and detail coefficients. The  analysis formulas (\ref{eq:haarsynthesis}) tell us that we can compute the labels of a node from the approximation coefficient of its son nodes, while the reconstruction formulas (\ref{eq:haarreconstruction}) tell us the reverse process is possible.  Since the Haar wavelet transform is invertible, we can equivalently represent the leaves of the tree (the original samples) using  all the detail coefficients $d_{j,k}$ in the non-leaf nodes and the approximation coefficient $a_{0,0}$ related to the root node.

\noindent This tree representation of the coefficients allows to clearly determine the dependency among them: a coefficient is determined by all and only the samples that are leaf nodes in the sub-tree rooted in itself. This idea has been used for example in \cite{budinich2017region} (in an adaptive setting) to reconstruct only a region of interest in an image while retaining some global information. 

It is possible to express this dependency explicitly: with a simple induction proof it can be seen that for $\ell =0,1,\ldots, L$, we have
\begin{align}\label{eq:haarexplicit}
  \begin{split}
a_{L-\ell,k} &= 2^{-\frac{\ell}{2}} \sum_{h=k2^\ell}^{(k+1)2^\ell - 1} a_{L,h} \\
\mbox{for } \ell > 0 \quad d_{L-\ell,k} &= 2^{-\frac{\ell}{2}} \left( \sum_{h=k2^\ell}^{(2k + 1)2^{\ell -1} - 1}a_{L,h} - \sum_{h=(2k + 1)2^{\ell -1}}^{(k+1)2^\ell - 1}a_{L,h} \right) \,\,,
\end{split}
\end{align}
i.e. we can write each approximation and detail wavelet coefficient as a linear combination of the samples $a_L$ themselves. Note that here $\ell$ is indicating the co-level, or equivalently the depth of the subtree rooted in the node.
If we define index sets
$$\mathcal{N}_{L - \ell,k} = \{k2^{\ell}, k2^{\ell} + 1,\ldots,(k+1)2^{\ell} - 1\}\,\,,$$
we can rewrite (\ref{eq:haarexplicit}) as
\begin{align}\label{eq:haarexplicit2}
  \begin{split}
a_{L-\ell,k} &= 2^{-\frac{\ell}{2}} \sum_{j \in \mathcal{N}_{L-\ell,k}} a_{L,j} \\
\mbox{for } \ell > 0 \quad d_{L-\ell,k} &= 2^{-\frac{\ell}{2}} \left( \sum_{j \in \mathcal{N}_{L - (\ell -1),2k}}a_{L,j} - \sum_{j \in \mathcal{N}_{L - (\ell -1),2k+1}}a_{L,j} \right) \,\,,
\end{split} 
\end{align}
which resemble formulas (\ref{D}), with the exception of a different normalization coefficient. This is an important distinction, arising from the fact that the Haar wavelet transform is deterministic and thus we know explicitly how the index sets $\mathcal{N}_{L - \ell,k}$ are made; in particular their cardinality depends only on the co-level $\ell$. In the definition of $D_{\ell,k}$ in (\ref{D}) instead we are weighing the sums over $\mathcal{N}_{\ell,2k}$ and $\mathcal{N}_{\ell,2k+1}$ with the reciprocal of their cardinalities, which will in general be different.

\begin{rem}\label{rem:mtap}
  Our approach to mimic the tree structure of the Haar wavelet filter bank to construct a dictionary could theoretically be extended to other, more complex wavelets. Suppose for example we were to choose a set of Daubechies 4 tap orthogonal filters: we would then need a clustering method that at each step partitions the data in $4$ subsets, e.g. 4-means (see also Remark \ref{rem:mary}). We could then use the weights given by the Daubechies filters to weight the representatives of the $4$ subsets of each node and compute the associated dictionary element. Though we didn't yet have time to experiment with this approach, we are interested to do so in the future. 
\end{rem}

\section{Algorithmic aspects}
\label{sec:alg}

In this Section we describe two implementations of our dictionary learning procedure described in Section \ref{sect:dict}. They differ for the order in which nodes in the tree are visited and the corresponding data points clustered: in the first variant we use a FIFO queue while in the second a Priority Queue. Both implementations allow us to build in one pass both  the Haar dictionary ${\mathbf D}^{H}$ and the leaves dictionary ${\mathbf D}^{L}$.

\paragraph{FIFO visit}
In pseudocode \ref{algo:fifo} we initialize a first-in-first-out (FIFO) queue and use it to visit the tree breadth-first. At each iteration we consider a node $\nu$ and proceed with adding its related dictionary elements into $\mathcal{D}^{L}$ and $\mathcal{D}^H$ only if
\begin{enumerate}
\item the node cardinality  \(|\nu|\) is above a threshold \texttt{mincard},
\item the value of the clustering minimization function $\mathcal{F}$ for the proposed clustering is above a threshold $\epsilon$.
\end{enumerate}
The function $\mathcal{F}$ on line \ref{algo:fifo:f} depends on the chosen clustering method: for 2-means  for example it will be the within cluster sum of squares. As mentioned in Section \ref{sect:dict}, the choice of representatives on line \ref{algo:fifo:representatives} depends on the clustering method.

In this variant we do not have direct control over the cardinality $K$ of the produced dictionary, we simply know that it will be a decreasing function of $\epsilon$. On the other hand we have the certainty that the final clusters will be very small: they either must have fewer than \texttt{mincard} elements or, when partitioned further, give a value of the clustering minimization function below $\epsilon$. This means that the clustering procedure gives some sort of adaptive resolution of the space: the tree branches go deeper where the data is more spread out, and in any case they go deep enough so that in all regions of the data space the final clusters have similar cardinality.

\begin{algorithm}[H]
\caption{FIFO based dictionary learning}
\label{algo:fifo}
\begin{algorithmic}[1]
\Require Training data ${\mathcal S} = \{Y_1,\ldots,Y_N\}$,  clustering procedure \textbf{ClusteringMethod}, $ \texttt{mincard} $, $ \epsilon $
\Ensure Haar dictionary $ \mathbf{D}^H$ and leaves dictionary $\mathbf{D}^L$ 
\State Initialize  $\mathbf{D}^H = \mathbf{D}^L :=  \{A_r \}$
\State Initialize \texttt{tovisit} =  FIFO
\State \texttt{tovisit}.put(\( r \))
\While{\(\texttt{tovisit} \) is not empty} \label{algo:fifo:whileline}
\State \( \nu \) = \texttt{tovisit}\mbox{.get}()
\If{\(|\nu|\leq \texttt{mincard}\)}
\State \textbf{goto} \ref{algo:fifo:whileline}
\EndIf
\State \(\mathcal{N}_{\nu_0} , \mathcal{N}_{\nu_1}  = \) \textbf{ClusteringMethod}\((\mathcal{N}_{\nu} )\)
\If{\(\mathcal{F}(\mathcal{N}_{\nu_0} , \mathcal{N}_{\nu_1}) > \epsilon\)} \label{algo:fifo:f}
\State Choose representatives $A_{\nu_{0}}$ and  $A_{\nu_{1}}$ \label{algo:fifo:representatives}
\If{\( \nu \neq r\)}
\State Remove \(A_{\nu} \) from  \( \mathbf{D}^L \)
\EndIf
\State Add edges \((\nu,\nu_{0})  \) and \( (\nu,\nu_{1}) \) to \( E \)
\State Add $A_{\nu_{0}}, A_{\nu_{1}}$ to $\mathbf{D}^L$ and $D_{\nu}$ to \( \mathbf{D}^{H} \)
\State \( \texttt{tovisit}\mbox{.put}(\nu_0) \)
\State \( \texttt{tovisit}\mbox{.put}(\nu_1) \)
\EndIf
\EndWhile
\end{algorithmic}
\end{algorithm}

\paragraph{Priority Queue visit}
In pseudocode \ref{algo:pq} we initialize a Priority queue, for which we'll use the variance of the nodes as key. This means that when we call the .get() on line \ref{algo:pq:get} we always receive the node representing the portion of data with highest variance, thus giving priority in the tree visit to those regions of the data space where the data is more spread out. At each iteration we consider a node $\nu$ and proceed with adding its related dictionary elements into $\mathcal{D}^{L}$ and $\mathcal{D}^H$ only if
\begin{enumerate}
\item the node cardinality $|\nu|$ is above a threshold \texttt{mincard},
\item the number of branchings already occurred is not greater than $K-1$.
\end{enumerate}
This means that, if the sample set is large enough, exactly $K-1$ branchings will occur, and thus  $\mathcal{D}^{L}$ and $\mathcal{D}^H$  will consist of $K$ dictionary elements.  In our tests we will always use this priority queue variant because of the convenience of setting the dictionary cardinality $K$.

In our tests, the two types of visit didn't produce any substantial difference in the generated dictionaries. The core of the method is the same, the only thing that changes is the order in which the nodes are visited. For applications we would suggest in general to use the FIFO queue, because of the lower operational cost of this data structure. However, when one wants to stop the visit after a fixed number of branchings (for example to obtain a dictionary of a specific cardinality $K$) then the Priority queue variant makes more sense, since regions of the data space that have a higher variance are given priority.

\begin{algorithm}[H]
\caption{Priority queue based dictionary learning}
\label{algo:pq}
\begin{algorithmic}[1]
\Require Training data ${\mathcal S} = \{Y_1,\ldots,Y_N\}$,  clustering procedure \textbf{ClusteringMethod}, $ \texttt{mincard} $, dictionary cardinality $ K \geq 2$ 
\Ensure Haar dictionary $ \mathbf{D}^H$ and leaves dictionary $\mathbf{D}^L$ 
\State Initialize  $\mathbf{D}^H = \mathbf{D}^L :=  \{A_r \}$ 
\State Initialize \texttt{tovisit} = PriorityQueue 
\State Put $r$ in \texttt{tovisit} with key $1$
\State nbranchings = 0
\While{\(\texttt{tovisit} \) is not empty} \label{algo:pq:whileline}
\State \( \nu \) = \texttt{tovisit}\mbox{.get}() \label{algo:pq:get}\Comment{Returns node with highest variance}
\If{\(|\nu|\leq \texttt{mincard}\)}
\State \textbf{goto} \ref{algo:pq:whileline}
\EndIf
\If{nbranchings \(\leq K - 1\)}
\State \(\mathcal{N}_{\nu_0} , \mathcal{N}_{\nu_1}  = \) \textbf{ClusteringMethod}\((\mathcal{N}_{\nu} )\)
\State Choose representatives $A_{\nu_{0}}$ and  $A_{\nu_{1}}$ \label{algo:pq:representatives}
\State nbranchings = nbranchings + 1
\If{\( \nu \neq r\)}
\State Remove \(A_{\nu} \) from  \( \mathbf{D}^L \)
\EndIf
\State Add edges \((\nu,\nu_{0})  \) and \( (\nu,\nu_{1}) \) to \( E \)
\State Add $A_{\nu_{0}}, A_{\nu_{1}}$ to $\mathbf{D}^L$ and $D_{\nu}$ to \( \mathbf{D}^{H} \)
\State Put $\nu_{0}$ in $\texttt{tovisit}$ with key $\mathrm{Var}[\mathcal{N}_{\nu_0}]$
\State Put $\nu_{1}$ in $\texttt{tovisit}$ with key $\mathrm{Var}[\mathcal{N}_{\nu_1}]$
\EndIf
\EndWhile
\end{algorithmic}
\end{algorithm}

The computational complexity of both variants essentially  depends on the clustering procedure used and on the number of branchings done (i.e., the number of nodes in the tree). Denoting with $\tilde{\mathcal N}$ the leaf nodes of the tree and supposing  that we use the 
2-means clustering by computing $\mathcal{I}$ iterations of Lloyd's algorithm, for each non-leaf node $\nu \in {\mathcal N} \setminus \tilde{\mathcal N}$ we require $\mathcal{O} (|\mathcal{S}_{\nu}| n \mathcal{I})$ elementary operations for the clustering  and $\mathcal{O}(|\mathcal{S}_{\nu}| n)$ for computing the associated dictionary element.  Thus in this case the total computational cost is
\begin{align}
\label{eq:gencompcost}
\sum_{\nu \in N \setminus \tilde{N}}  {\mathcal O}(|\mathcal{S}_{\nu}| n \mathcal{I})  \leq  {\mathcal O}(K N n \mathcal{I})  \,\,. 
\end{align}

\begin{rem}
Our method can be adapted to online dictionary learning (see for example \cite{mairal2009online} and \cite{lu2013online}). In this scenario one wishes to update the dictionary based on new incoming learning data. If we suppose that the structure of the hierarchical clustering remains unchanged even with the addition of the new training data, it is sufficient to assign each new data point to the cluster corresponding to one of the leaves and then travel on the tree from these leaf nodes up to the root node. Only the dictionary atoms associated to nodes so encountered (i.e. the ancestors of the leaf nodes containing the new data points) will be affected. 

\noindent If the new incoming data presents very different features than the original training data, then it would produce substantially different clusters and the hypotheses of the tree not changing would become unrealistic. One could identify the regions of the data space that are changing and recompute only the corresponding subtrees.

\noindent Finally, our method can be used to produce, with no further computational costs, subdictionaries adapted to only a portion of the data. This can be done by simply selecting an appropriate subtree and the dictionary atoms associated to it. We could apply this method for example to accelerate the sparse coding procedure, by first assigning a sample to one of the leaf clusters and then selecting a subtree containing this leaf, whose associated dictionary would be used for sparse coding. One could thus regulate with a parameter the trade-off between speed and accuracy of the sparse coding method: a more shallow tree would correspond to a smaller dictionary and thus faster computation times.  
\end{rem}

\section{Results}
\label{sec:num}
In this Section we will carry out natural image reconstruction tasks using K-SVD and particular variants of our method. We will compare computation times and the quality of the reconstructions using the HaarPSI index (\cite{reisenhofer2018haar}). The HaarPSI of two images is a real number in \((0,1]\) indicating the visual similarity of two images, where \(1\) means the two images are the same and a lower number indicates higher distortion. We choose this index because we are testing reconstruction of natural images and the HaarPSI has the best correlation with human subjective quality assessment.  The implementation of our method was done in python\footnote{available at \url{https://github.com/nareto/haardict}} while for K-SVD we used the KSVD-box Matlab software\footnote{avaiable at \href{http://www.cs.technion.ac.il/\~ronrubin/software.html}{http://www.cs.technion.ac.il/\textasciitilde{}ronrubin/software.html}}. All the numerical tests were run on a MacBook Pro Mid 2012 with an Intel Ivy Bridge i5 2.5Ghz CPU. The exact code used to produce the results in this section can be found in the \texttt{batch\_tests.py} file in the git repository. 

\begin{figure}[htbp]
\centerline{\includegraphics[scale=0.15]{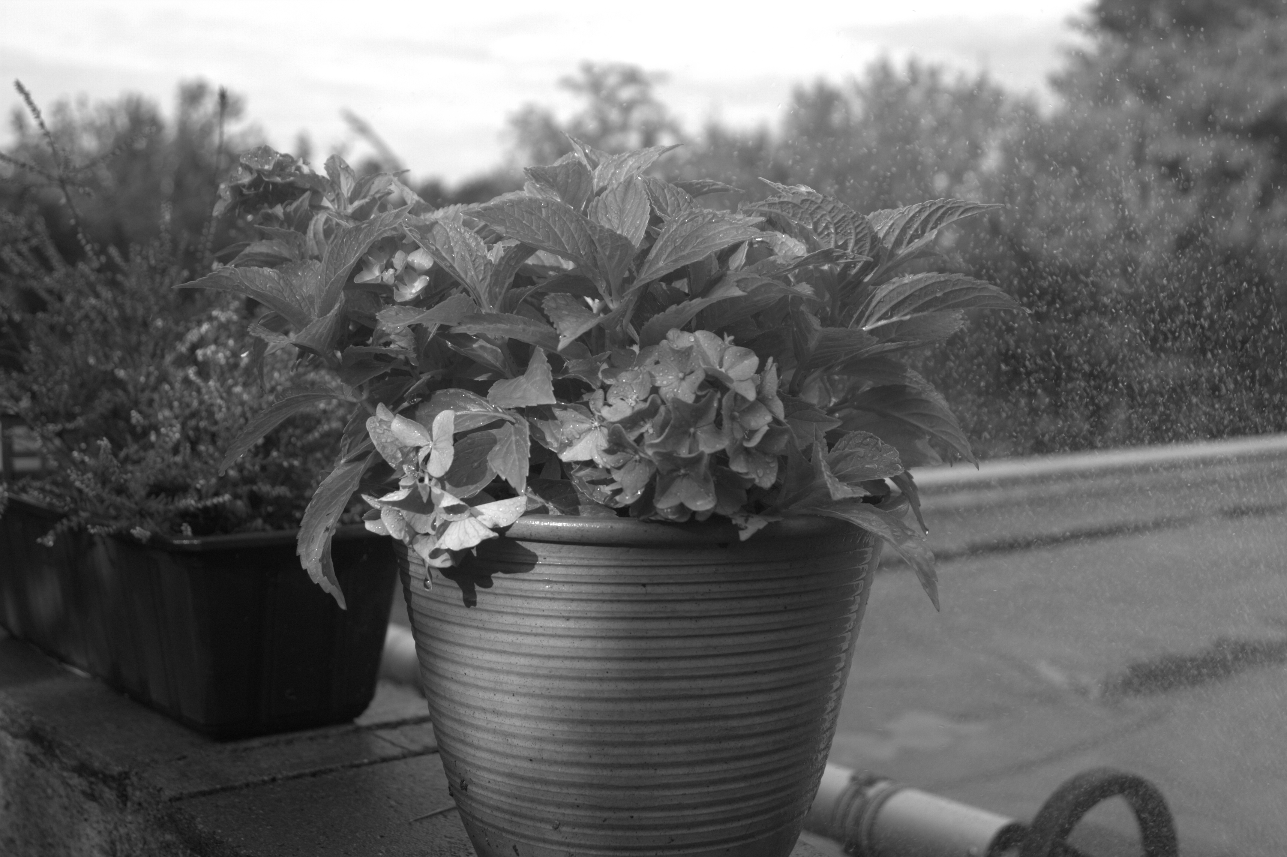}}
\caption{\texttt{flowers\_pool} image\label{flowerspool}}
\end{figure}

Using patches extracted from the \texttt{flowers\_pool} image (Figure \ref{flowerspool}) as training data, we computed the dictionaries with various methods and used OMP to reconstruct patches from the same image. While we randomly extracted patches from the set of overlapping patches in the image, we reconstructed non-overlapping patches due to time inefficiency of OMP when dealing with a large number of data points. We ran the test with different values of patch number, patch size, clustering method and reconstruction sparsity.  For clustering we used the classical K-means method with \(K=2\), the K-maxoids method (\cite{bauckhage2015k}) with \(K=2\) and the Spectral Clustering method (\cite{shi2000normalized}) with different data graphs. The K-maxoids is slightly slower than K-means but offers as class representatives some particular patches in the data-set as opposed to the cluster centroids. In our case we hypothesized this would be an advantage, since it would give us dictionary patches that are more sharp and less blurry, which will be summed in linear combinations anyways by OMP. Spectral Clustering relies on what we call the data similarity graph, a complete graph with vertices given by patches and edge weights given by their similarity under some measure. For the latter we used the Frobenius norm, the aforementioned HaarPSI and the Earth Movers' Distance (see for example \cite{rubner2000earth}). Because of the \(O(N^2)\) computations of such similarity measure required spectral clustering is much slower and unusable for larger patch sizes and quantity; we thus restricted the computation of this clustering to simpler cases. In all cases we set the dictionary cardinality to be \(50\%\) bigger than the dimension of the vectorized patches, i.e. for \(8 \times 8\) patches we computed dictionaries with \(96\) atoms. 

\begin{figure}[htbp]
\vspace*{-10mm}

\centerline{\includegraphics[scale=0.35]{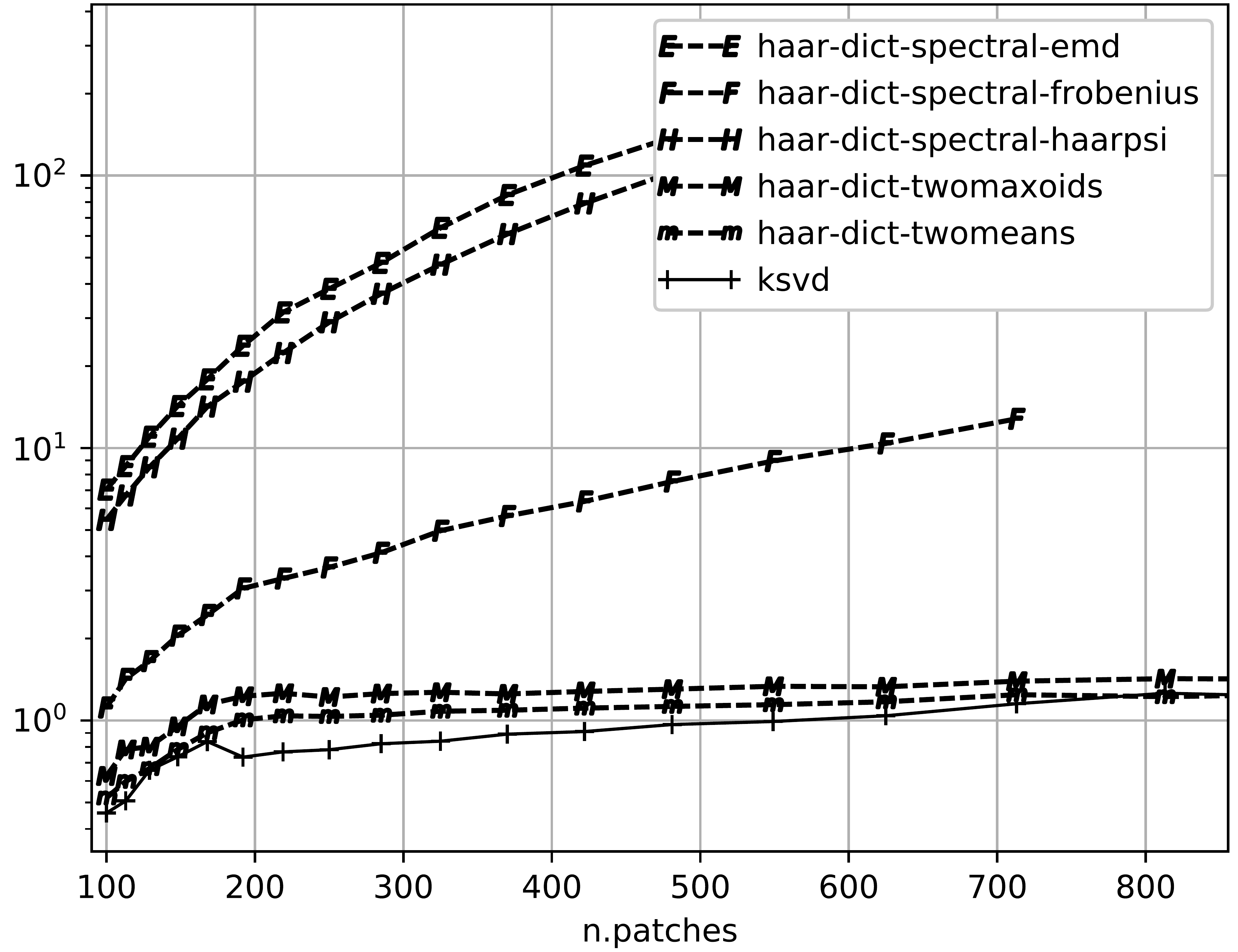}}

\centerline{ \includegraphics[scale=0.35]{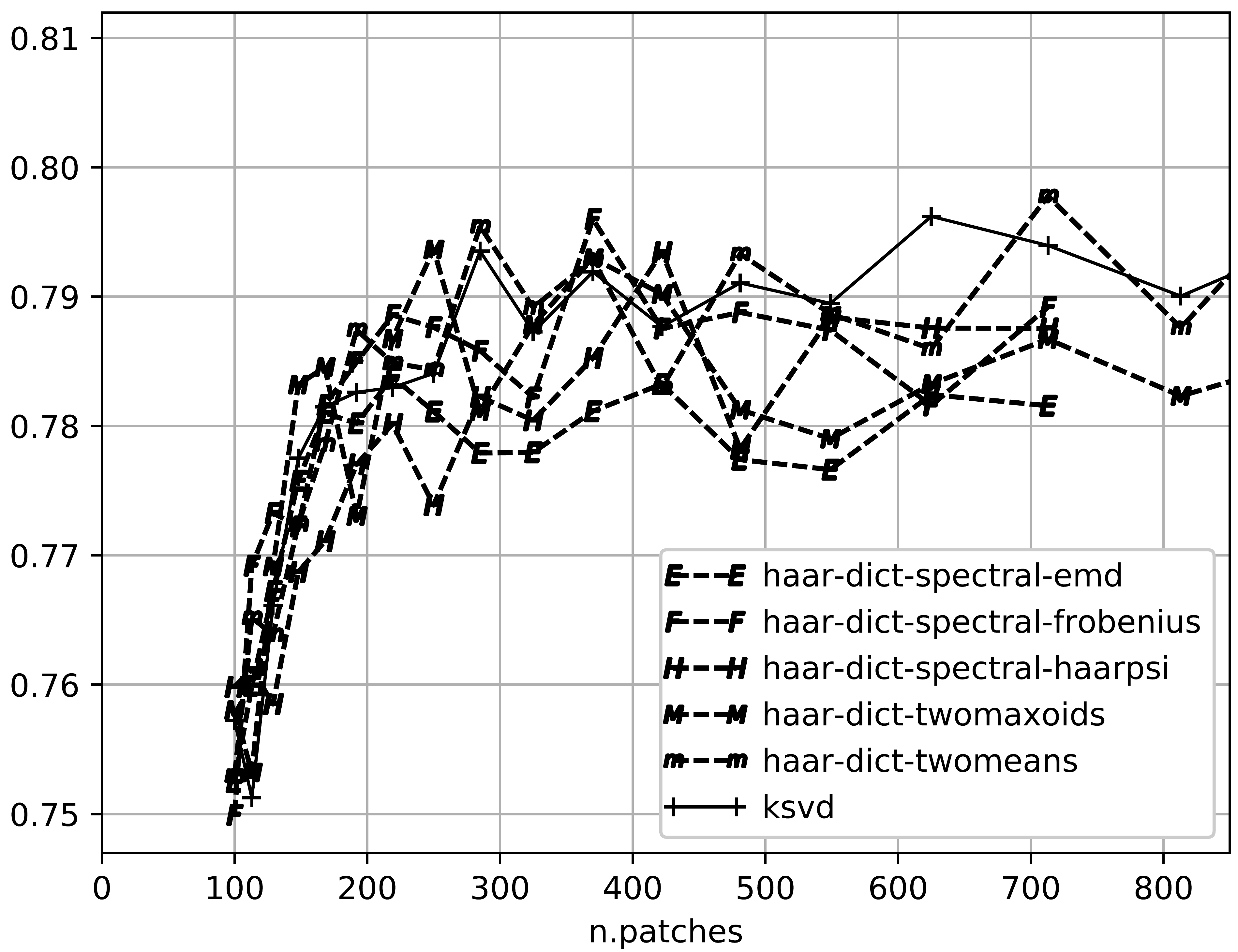}}
\caption{Top: Times in seconds (in logarithmic scale) required to learn the dictionaries as a function of the number of \( 8 \times 8 \) patches used for training. Bottom: HaarPSI values of the reconstructed images with sparsity \( 5 \) as a function of the number of  patches used for training.
\label{spectimes}}
\end{figure}

In Figure \ref{spectimes} (top) the computation times to learn various dictionaries are shown, in logarithmic scale. It can be clearly seen that spectral clustering performs much worse than 2-means or 2-maxoids, especially when using HaarPSI or Earth Mover's Distance as similarity measure. The reconstruction HaarPSI values (shown in Figure \ref{spectimes}  (bottom)) are only in certain cases better than other methods. Overall we consider spectral clustering's computation times prohibitive for anything but very small number of data points, and we thus excluded it from further tests.

\begin{figure}[htbp]
\vspace*{-10mm}

  \centerline{
    \includegraphics[scale=0.9]{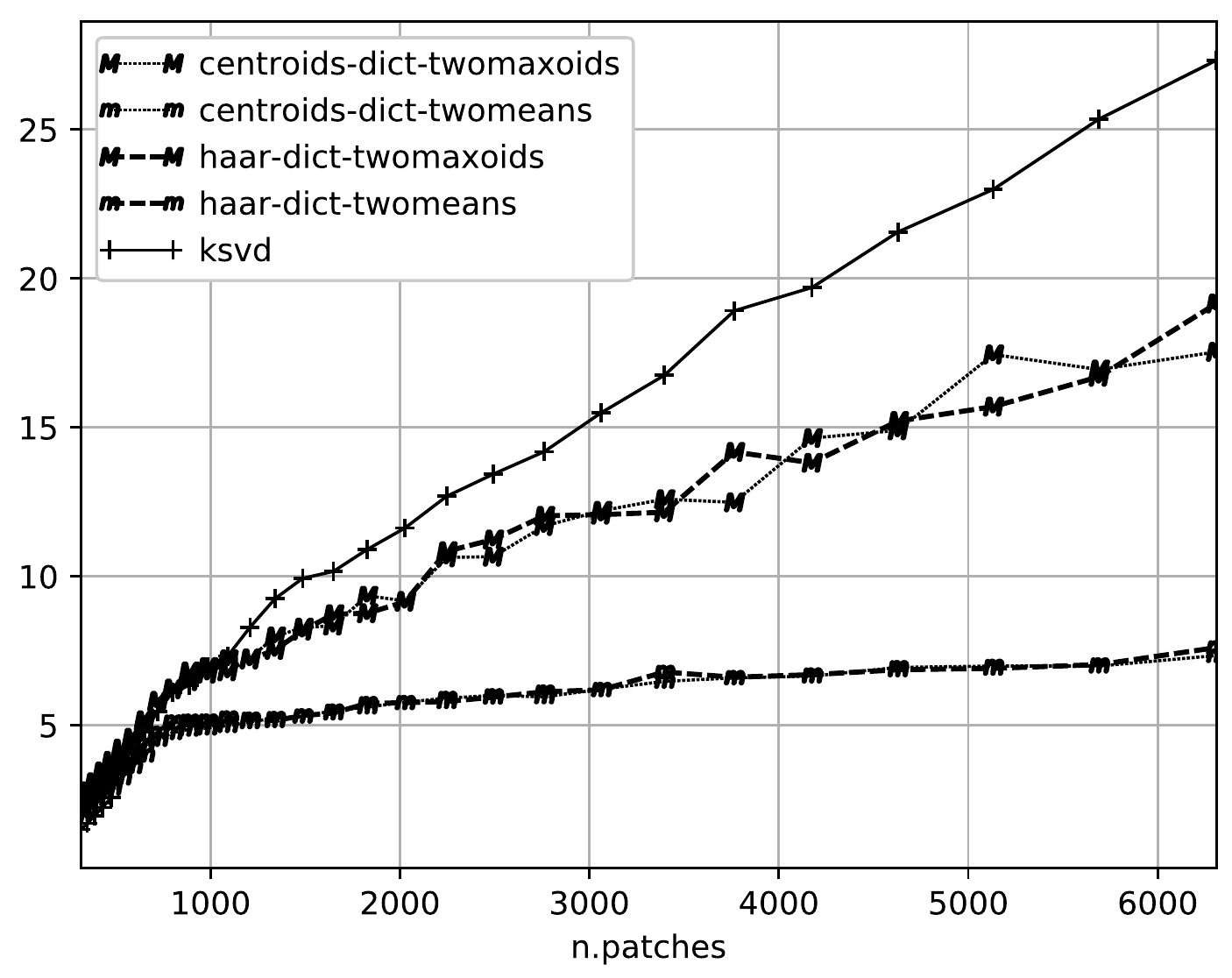}}
\centerline{    \includegraphics[scale=0.9]{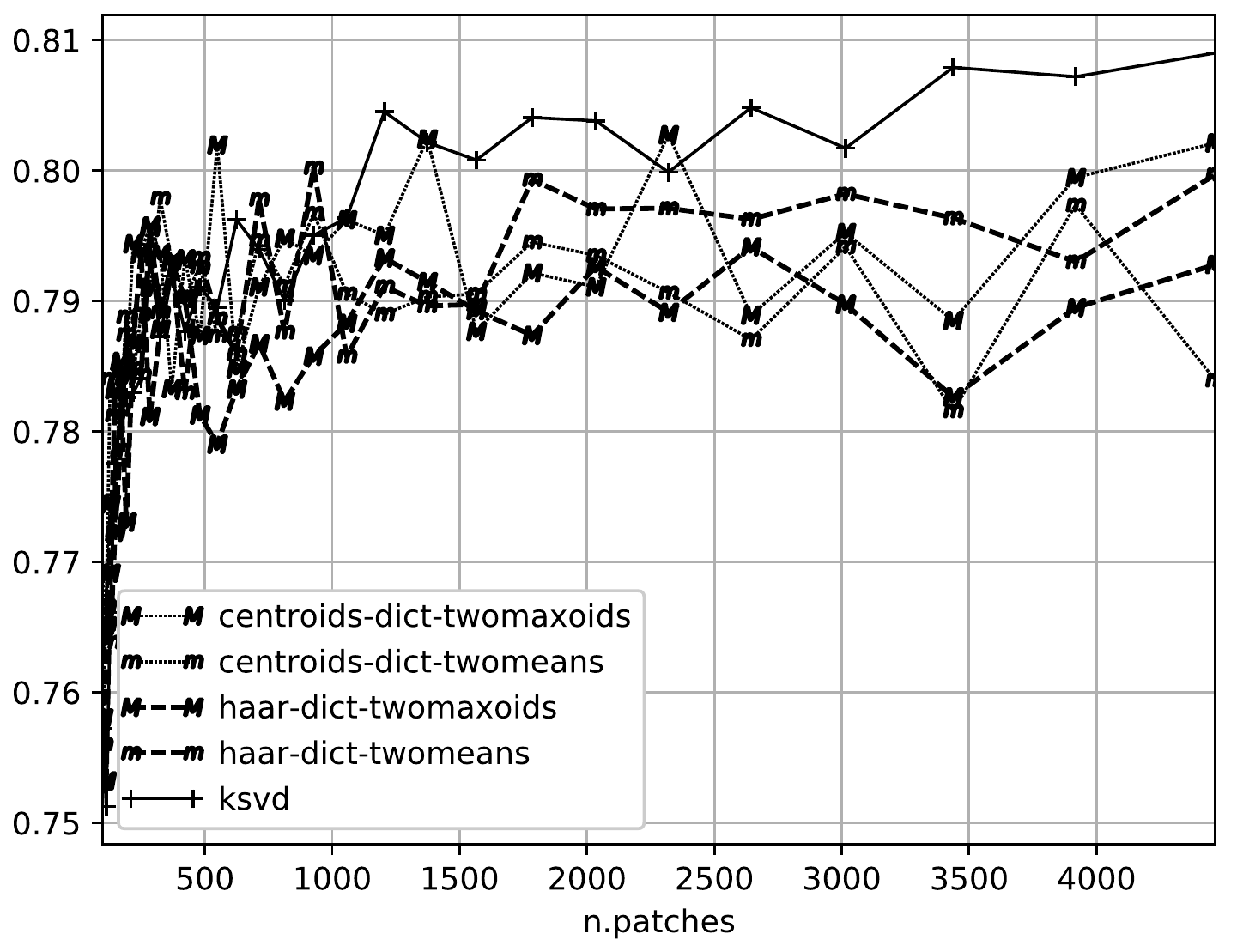}}
\caption{Top: Times in seconds required to learn the dictionaries as a function of the number of \( 16 \times 16 \) patches used for training. Bottom: HaarPSI values of the reconstructed images with sparsity \( 4 \) as a function of the number of  patches used for training. \label{times88}}
\end{figure}

\noindent In Figure \ref{times88}  (top) we plot the computation times required for learning dictionaries trained on different number of \(8 \times 8\) patches. In Figure \ref{times88} (bottom) instead we plot the HaarPSI values of the reconstructed images (with sparsity \(4\)) from these dictionaries.  The learning times clearly show the better performance of our method, especially when using 2-means clustering. K-SVD still gives better quality reconstructions though, followed by the Haar-dictionary with 2-means clustering and the leaves dictionary with 2-maxoids clustering. 

\begin{figure}[htbp]
\begin{subfigure}[b]{0.5\textwidth}
\includegraphics[scale=0.5,trim= 30px 15px 30px 30px, clip]{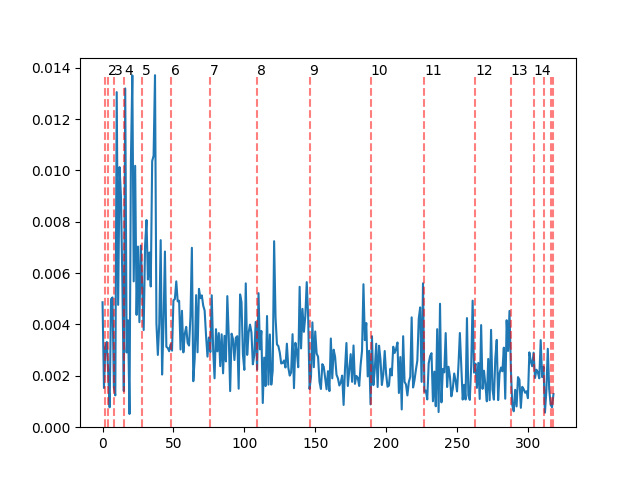}
\caption{Haar-dictionary}
\end{subfigure}\hspace{0.4em}%
\begin{subfigure}[b]{0.5\textwidth}
\includegraphics[scale=0.5, trim= 30px 15px 30px 30px,clip]{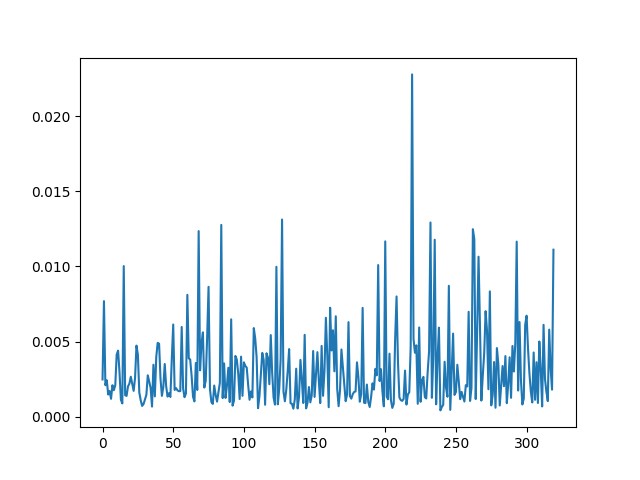}%
\caption{K-SVD dictionary}
\end{subfigure}
\caption{Values of \( \eta_k \) defined in \eqref{eq:etak} for the Haar-dictionary (with 2-means clustering and FIFO tree visit with $\epsilon=1$) and K-SVD dictionary with \( 320 \) elements computed on the \( 32\times 32 \) patches of the \texttt{flowers-pool} image when used for the reconstruction of this same image. The red lines and the numbers on the top indicate the level in the tree of the corresponding dictionary elements.}\label{fig:atomsprob} 
\end{figure}

We observed that our Haar-dictionary captures some structure that is not present in the K-SVD atoms: dictionary atoms associated to nodes at smaller levels in the tree (i.e. closer to the root node) are used with larger coefficients by OMP. To see this we consider the solution \( {\mathbf X}\) (in the notation of \eqref{eq:sparsecodingprob}) proposed by OMP, we sum its rows in absolute values and associate these numbers to the corresponding dictionary atoms; we define
\begin{align}
\label{eq:etak}
 \eta_k := \sum_{j=1}^N \left| X_{kj}\right|, \qquad  k =1,\ldots, K \, .
\end{align}
The number \(\eta_k\) gives us a measure of how important the dictionary atom \(k\) is, in the sense that it is more used in the sparse linear combinations of the reconstructed patches. In Figure \ref{fig:atomsprob} we represent the vectors \(\eta\) for the Haar-dictionary (with 2-means clustering) and the K-SVD dictionary: it can be seen in both plots (and this is mostly the case in all the tests we've conducted) that there are few atoms that are used very frequently in the reconstruction and other atoms that are used with far less frequency. The difference however is that the plot for the Haar-dictionary presents a decreasing trend: atoms that are computed earlier are more used by OMP. Since in this case the FIFO tree visit strategy was used, these atoms correspond to the first levels of the tree: this means that the atoms that OMP uses the most in the sparse coding procedure are given by the differences between representatives of large clusters, i.e., they distinguish between features of the data at a very coarse level.  

\noindent This property could be used to obtain a sub-dictionary with similar reconstruction power by limiting the tree-depth; this would accelerate OMP. 
We remark that the atoms in this sub-dictionary associated to nodes closer to the root node would be stable to variations in the data-set given for example by noise, since they represent coarse-level features in the data.

\begin{figure}[htbp]
\begin{center}
\begin{subfigure}[b]{\textwidth}
\centering{\includegraphics[scale=2.2,trim= 30px 260px 100px 35px, clip]{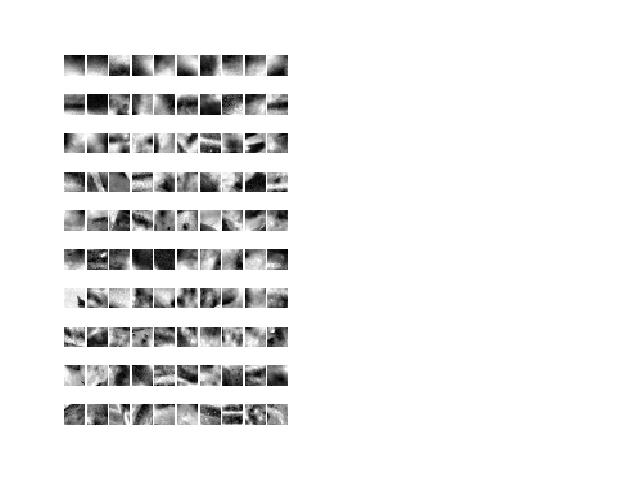}}
\caption{\( 2 \)-means Haar-dictionary}
\end{subfigure} 
\begin{subfigure}[b]{\textwidth}
\centering{\includegraphics[scale=2.2,trim= 30px 260px 100px 35px, clip]{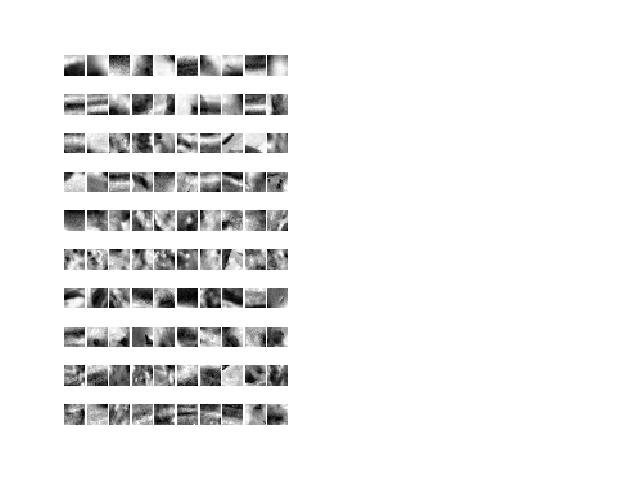}}
\caption{Spectral Clustering (with HaarPSI similarity measure) Haar-dictionary}
\end{subfigure} 
\begin{subfigure}[b]{\textwidth}
\centering{\includegraphics[scale=2.2,trim= 30px 260px 100px 35px, clip]{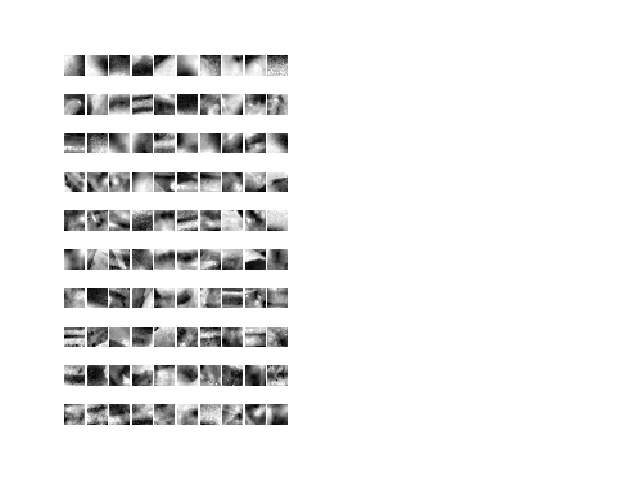}}
\caption{\( 2 \)-maxoids Haar-dictionary}
\end{subfigure} 
\begin{subfigure}[b]{\textwidth}
\centering{\includegraphics[scale=2.2,trim= 30px 260px 100px 35px, clip]{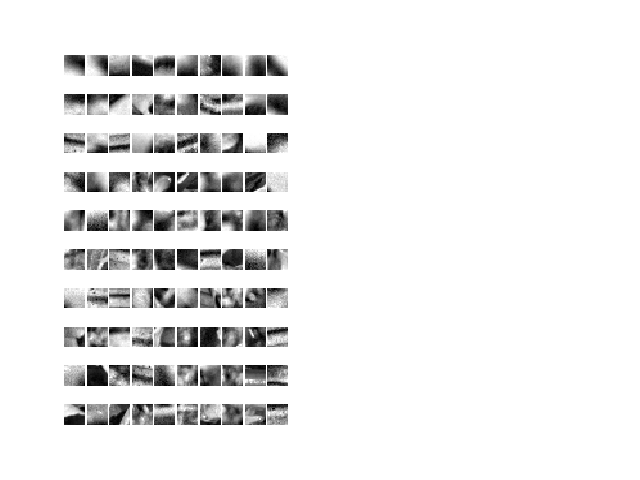}}
\caption{\texttt{K-SVD} dictionary}
\end{subfigure} 
\caption{\( 20 \) dictionary atoms with highest \( \eta_k \) values (when reconstructing with sparsity \( 5 \)) for various dictionaries. All the dictionaries were trained on \( 500 \) \( 16 \times 16 \) patches extracted from the \texttt{flowers\_pool} image.\label{dictpatches}}
\end{center}
\end{figure}

Finally in Figure \ref{dictpatches} we show the most used (i.e. ordered by decreasing values of $\eta_k$) dictionary patches for various dictionaries. It can be seen that when using \(2\)-means our dictionary produces very smoothed out patches; this is due to the Haar-dictionary elements being difference of centroids of sibling clusters. The patches obtained instead using the \(2\)-maxoids clustering have, as expected, sharper edges.

\section{Conclusion}
We proposed a new scheme for dictionary learning which is fast and flexible. By using one from the many available clustering methods we derive an adaptive multiscale analysis of the data, which generalizes the tree structure of the classical Haar wavelet. Thanks to this common structure we are able to define the dictionary atoms analogously to the Haar wavelet coefficients in linear computation time. Thus the multi-scale structure reflects itself in the dictionary, with atoms corresponding to higher nodes in the tree being used with bigger coefficients by OMP in the final sparse coding step. In image reconstruction tasks we achieve similar quality as K-SVD and, when using K-means for the clustering, much faster computation times.

\subsection*{Acknowledgement}
The authors gratefully acknowledge support by the German Research Foundation in the framework of the RTG 2088.

\small
\bibliographystyle{apalike}

\bibliography{biblio-v4}{}

\begin{thebibliography}{}

\bibitem[Aharon et~al., 2006]{aharon2006img}
Aharon, M., Elad, M., and Bruckstein, A. (2006).
\newblock K-svd: An algorithm for designing overcomplete dictionaries for
  sparse representation.
\newblock {\em IEEE Trans. Signal Process.}, 54(11):4311--4322.

\bibitem[Bauckhage and Sifa, 2015]{bauckhage2015k}
Bauckhage, C. and Sifa, R. (2015).
\newblock k-{m}axoids clustering.
\newblock In {\em LWA}, pages 133--144.

\bibitem[Beck and Teboulle, 2009]{beck09}
Beck, A. and Teboulle, M. (2009).
\newblock A fast iterative shrinkage-thresholding algorithm for linear inverse
  problems.
\newblock {\em SIAM J. Imaging Sci.}, 2(1):183--202.

\bibitem[Blumensath and Davies, 2008]{blumensath2008iterative}
Blumensath, T. and Davies, M.~E. (2008).
\newblock Iterative thresholding for sparse approximations.
\newblock {\em J. Fourier Anal. Appl.}, 14(5-6):629--654.

\bibitem[Budinich, 2017]{budinich2017region}
Budinich, R. (2017).
\newblock A region-based easy-path wavelet transform for sparse image
  representation.
\newblock {\em Int. J. Wavelets Multiresolut. Inf. Process.}, 15(05):1750045.

\bibitem[Budinich, 2018]{budinich18}
Budinich, R. (2018).
\newblock {\em Adaptive Multiscale Methods for Sparse Image Representation and
  Dictionary Learning}.
\newblock PhD thesis, University of G\"ottingen.

\bibitem[Cai et~al., 2014]{cai14}
Cai, J., Ji, H., Shen, Z., and Ye, G. (2014).
\newblock Data-driven tight frame construction and image denoising.
\newblock {\em Appl. Comput. Harmon. Anal.}, 37(1):89--105.

\bibitem[Chambolle and Pock, 2011]{chambolle11}
Chambolle, A. and Pock, T. (2011).
\newblock A first-order primal-dual algorithm for convex problems with
  applications to imaging.
\newblock {\em J. Math. Imaging Vis.}, 40(1):120--145.

\bibitem[Damelin and Miller~Jr, 2012]{damelin2012mathematics}
Damelin, S.~B. and Miller~Jr, W. (2012).
\newblock {\em The {M}athematics of {S}ignal {P}rocessing}, volume~48.
\newblock Cambridge University Press.

\bibitem[Davies et~al., 1997]{davies97}
Davies, G., Mallat, S., and Avellaneda, M. (1997).
\newblock Adaptive greedy approximations.
\newblock {\em Constr. Approx.}, 13(1):57--98.

\bibitem[Dragotti and Lu, 2014]{dragotti2014sparse}
Dragotti, P.~L. and Lu, Y.~M. (2014).
\newblock On sparse representation in {Fourier} and local bases.
\newblock {\em IEEE Trans. Inf. Theory}, 60(12):7888--7899.

\bibitem[Eksioglu and Bayir, 2014]{eksio2014}
Eksioglu, E.~M. and Bayir, O. (2014).
\newblock K-svd meets transform learning: Transform k-svd.
\newblock {\em IEEE Signal Process. Letters}, 21(3):347--351.

\bibitem[Elad, 2010]{elad2010sparse}
Elad, M. (2010).
\newblock {\em Sparse and Redundant Representations: From Theory to
  Applications in Signal and Image Processing}.
\newblock Springer Publishing Company, Incorporated, 1st edition.

\bibitem[Jenatton et~al., 2011]{jenat2011}
Jenatton, R., Mairal, J., Obozinski, G., and Bach, F. (2011).
\newblock Proximal methods for sparse hierarchical dictionary learning.
\newblock {\em Journal of Machine Learning Research}, 12:2297--2334.

\bibitem[Kong et~al., 2005]{kong2005generalized}
Kong, H., Wang, L., Teoh, E.~K., Li, X., Wang, J.-G., and Venkateswarlu, R.
  (2005).
\newblock Generalized 2d principal component analysis for face image
  representation and recognition.
\newblock {\em Neural Networks}, 18(5):585--594.

\bibitem[Liu et~al., 2018]{liu2018sparse}
Liu, L., Ma, J., and Plonka, G. (2018).
\newblock Sparse graph-regularized dictionary learning for suppressing random
  seismic noise.
\newblock {\em Geophysics}, 83(3):V215--V231.

\bibitem[Liu et~al., 2017]{liu17}
Liu, L., Plonka, G., and Ma, J. (2017).
\newblock Seismic data interpolation and denoising by learning a tensor tight
  frame.
\newblock {\em Inverse Problems}, 33(10):105011.

\bibitem[Lu et~al., 2013]{lu2013online}
Lu, C., Shi, J., and Jia, J. (2013).
\newblock Online robust dictionary learning.
\newblock In {\em Proceedings of the IEEE Conference on Computer Vision and
  Pattern Recognition}, pages 415--422.

\bibitem[Mairal et~al., 2009]{mairal2009online}
Mairal, J., Bach, F., Ponce, J., and Sapiro, G. (2009).
\newblock Online dictionary learning for sparse coding.
\newblock In {\em Proceedings of the 26th annual international conference on
  machine learning}, pages 689--696. ACM.

\bibitem[Mairal et~al., 2010]{mairal2010}
Mairal, J., Bach, F., Ponce, J., and Sapiro, G. (2010).
\newblock Online learning for matrix factorization and sparse coding.
\newblock {\em Journal of Machine Learning Research}, 11:19--60.

\bibitem[Mallat, 2008]{mallat2008wavelet}
Mallat, S. (2008).
\newblock {\em A {W}avelet {T}our of {S}ignal {P}rocessing: {T}he {S}parse
  {W}ay}.
\newblock Academic Press.

\bibitem[Mazaheri et~al., 2013]{maza2013}
Mazaheri, J.~A., Guillemot, C., and Labit, C. (2013).
\newblock Learning a tree-structured dictionary for efficient image
  representation with adaptive sparse coding.
\newblock In {\em 2012 IEEE International Conference on Acoustics, Speech and
  Signal Processing (ICASSP)}, pages 1320--1324. IEEE.

\bibitem[Murtagh, 2007]{murtagh2007haar}
Murtagh, F. (2007).
\newblock The {H}aar wavelet transform of a dendrogram.
\newblock {\em J. Classification}, 24(1):3--32.

\bibitem[Natarajan, 1995]{natarajan1995sparse}
Natarajan, B.~K. (1995).
\newblock Sparse approximate solutions to linear systems.
\newblock {\em SIAM J. Comput.}, 24(2):227--234.

\bibitem[Nguyen et~al., 2012]{nguyen2012}
Nguyen, H.~V., Patel, V.~M., Nasrabadi, N.~M., and Chellappa, R. (2012).
\newblock Kernel dictionary learning.
\newblock In {\em 2012 IEEE International Conference on Acoustics, Speech and
  Signal Processing (ICASSP)}, pages 2021--2024. IEEE.

\bibitem[Ophir et~al., 2011]{ophir2011}
Ophir, B., Lustig, M., and Elad, M. (2011).
\newblock Multi-scale dictionary learning using wavelets.
\newblock {\em IEEE Journal of Selected Topics in Signal Processing},
  5(5):1014--1024.

\bibitem[Pati et~al., 1993]{pati1993orthogonal}
Pati, Y.~C., Rezaiifar, R., and Krishnaprasad, P.~S. (1993).
\newblock Orthogonal matching pursuit: Recursive function approximation with
  applications to wavelet decomposition.
\newblock In {\em Signals, Systems and Computers, 1993. 1993 Conference Record
  of The Twenty-Seventh Asilomar Conference on}, pages 40--44. IEEE.

\bibitem[Reisenhofer et~al., 2018]{reisenhofer2018haar}
Reisenhofer, R., Bosse, S., Kutyniok, G., and Wiegand, T. (2018).
\newblock A {Haar} wavelet-based perceptual similarity index for image quality
  assessment.
\newblock {\em Signal Process., Image Commun.}, 61:33--43.

\bibitem[Rubinstein et~al., 2013]{rubin2013}
Rubinstein, R., Peleg, T., and Elad, M. (2013).
\newblock Analysis k-svd: A dictionary-learning algorithm for the analysis
  sparse model.
\newblock {\em IEEE Trans. Signal Process.}, 61(3):661--677.

\bibitem[Rubner et~al., 2000]{rubner2000earth}
Rubner, Y., Tomasi, C., and Guibas, L.~J. (2000).
\newblock The earth mover's distance as a metric for image retrieval.
\newblock {\em Int. J. Comput. Vis.}, 40(2):99--121.

\bibitem[Shen et~al., 2015]{shen2015}
Shen, L., Huang, Q., Wang, S., Lin, Z., and Wu, E. (2015).
\newblock Multi-level discriminative dictionary learning with application to
  large scale image classification.
\newblock {\em IEEE Trans. Image Process.}, 24(10):3109--3123.

\bibitem[Shi and Malik, 2000]{shi2000normalized}
Shi, J. and Malik, J. (2000).
\newblock Normalized cuts and image segmentation.
\newblock {\em IEEE Trans. Pattern Anal. Mach. Intell.}, 22(8):888--905.

\bibitem[Sulam et~al., 2016]{sulam2016}
Sulam, J., Ophir, B., Zibulevsky, M., and Elad, M. (2016).
\newblock Trainlets: Dictionary learning in high dimensions.
\newblock {\em IEEE Trans. Signal Process.}, 64(12):3180--3193.

\bibitem[Tillmann, 2015]{tillmann2015computational}
Tillmann, A.~M. (2015).
\newblock On the computational intractability of exact and approximate
  dictionary learning.
\newblock {\em IEEE Signal Process. Lett.}, 22(1):45--49.

\bibitem[Tropp, 2004]{tropp04}
Tropp, J.~A. (2004).
\newblock Greed is good: Algorithmic results for sparse approximation.
\newblock {\em IEEE Trans. Inf. Theory}, 50(10):2231--2242.

\bibitem[Yan et~al., 2009]{yan2009fast}
Yan, D., Huang, L., and Jordan, M.~I. (2009).
\newblock Fast approximate spectral clustering.
\newblock In {\em Proceedings of the 15th ACM SIGKDD international conference
  on Knowledge discovery and data mining}, pages 907--916. ACM.

\bibitem[Yankelevsky and Elad, 2016]{yanke16}
Yankelevsky, Y. and Elad, M. (2016).
\newblock Dual graph regularized dictionary learning.
\newblock {\em IEEE Trans. Signal Inf. Process. Netw.}, 2(4):611--624.

\bibitem[Zeng et~al., 2015]{zeng2015dictionary}
Zeng, X., Bian, W., Liu, W., Shen, J., and Tao, D. (2015).
\newblock Dictionary pair learning on {G}rassmann manifolds for image
  denoising.
\newblock {\em IEEE Trans. Image Process.}, 24(11):4556--4569.

\end{thebibliography}
\end{document}